\pdfoutput=1

\documentclass[11pt]{article}

\usepackage[final]{acl}

\usepackage{times}
\usepackage{latexsym}
\usepackage{tabularx}
\usepackage{booktabs}
\usepackage{caption}
\usepackage{array}
\usepackage{multirow}
\usepackage{amsmath}
\usepackage[T1]{fontenc}

\usepackage[utf8]{inputenc}

\usepackage{microtype}
\usepackage{array} 
\usepackage{pifont} 
\usepackage{adjustbox}

\usepackage{inconsolata}

\usepackage{graphicx}
\title{Debate-to-Detect: Reformulating Misinformation Detection as a Real-World Debate with Large Language Models}

\author{
  \textbf{Chen Han}\textsuperscript{1,2}, 
  \textbf{Wenzhen Zheng}\textsuperscript{2}, 
  \textbf{Xijin Tang}\textsuperscript{1,2} \\
  \textsuperscript{1} School of Advanced Interdisciplinary Sciences, University of Chinese Academy of Sciences \\
  \textsuperscript{2} State Key Laboratory of Mathematical Sciences, Academy of Mathematics and Systems Science, \\  Chinese Academy of Sciences \\
  \texttt{\{hanchen23, zhengwenzhen21\}@mails.ucas.ac.cn}, \texttt{xjtang@iss.ac.cn}
}

\begin{document}
\maketitle
\begin{abstract}
The proliferation of misinformation in digital platforms reveals the limitations of traditional detection methods, which mostly rely on static classification and fail to capture the intricate process of real-world fact-checking. Despite advancements in Large Language Models (LLMs) that enhance automated reasoning, their application to misinformation detection remains hindered by issues of logical inconsistency and superficial verification. 
Inspired by the idea that "Truth Becomes Clearer Through Debate", we introduce Debate-to-Detect (D2D), a novel Multi-Agent Debate (MAD) framework that reformulates misinformation detection as a structured adversarial debate. Based on fact-checking workflows, D2D assigns domain-specific profiles to each agent and orchestrates a five-stage debate process, including Opening Statement, Rebuttal, Free Debate, Closing Statement, and Judgment. To transcend traditional binary classification, D2D introduces a multi-dimensional evaluation mechanism that assesses each claim across five distinct dimensions: Factuality, Source Reliability, Reasoning Quality, Clarity, and Ethics. Experiments with GPT-4o on two fakenews datasets demonstrate significant improvements over baseline methods, and the case study highlight D2D's capability to iteratively refine evidence while improving decision transparency, representing a substantial advancement towards robust and interpretable misinformation detection. The code will be released publicly after the official publication.

\end{abstract}

\section{Introduction}
The modern information landscape is flooded with content that may be linguistically fluent but factually misleading, ranging from political rumors to health misinformation \citep{aimeur2023fakenews,spampatti2024psychological,saha-srihari-2024-integrating}. While large language models (LLMs) such as GPT‑4o show advanced capabilities on many reasoning benchmarks \citep{madaan2023selfrefine,liang-etal-2024-encouraging}, their reliability in evaluating the factuality of real-world news remains limited \citep{gou2024critic,ma-etal-2024-fake}. When exposed to misleading narratives, LLMs often “take the text at face value,” leading to overconfident yet inaccurate judgments \citep{He2023}. Such challenges can be attributed to their reliance on surface-level linguistic patterns rather than deep contextual understanding, leading to not only misinformation detection failures but also potential amplification \citep{pan-etal-2023-risk,10.1145/3626772.3661377}.

\begin{figure}[t]
  \includegraphics[width=\columnwidth]{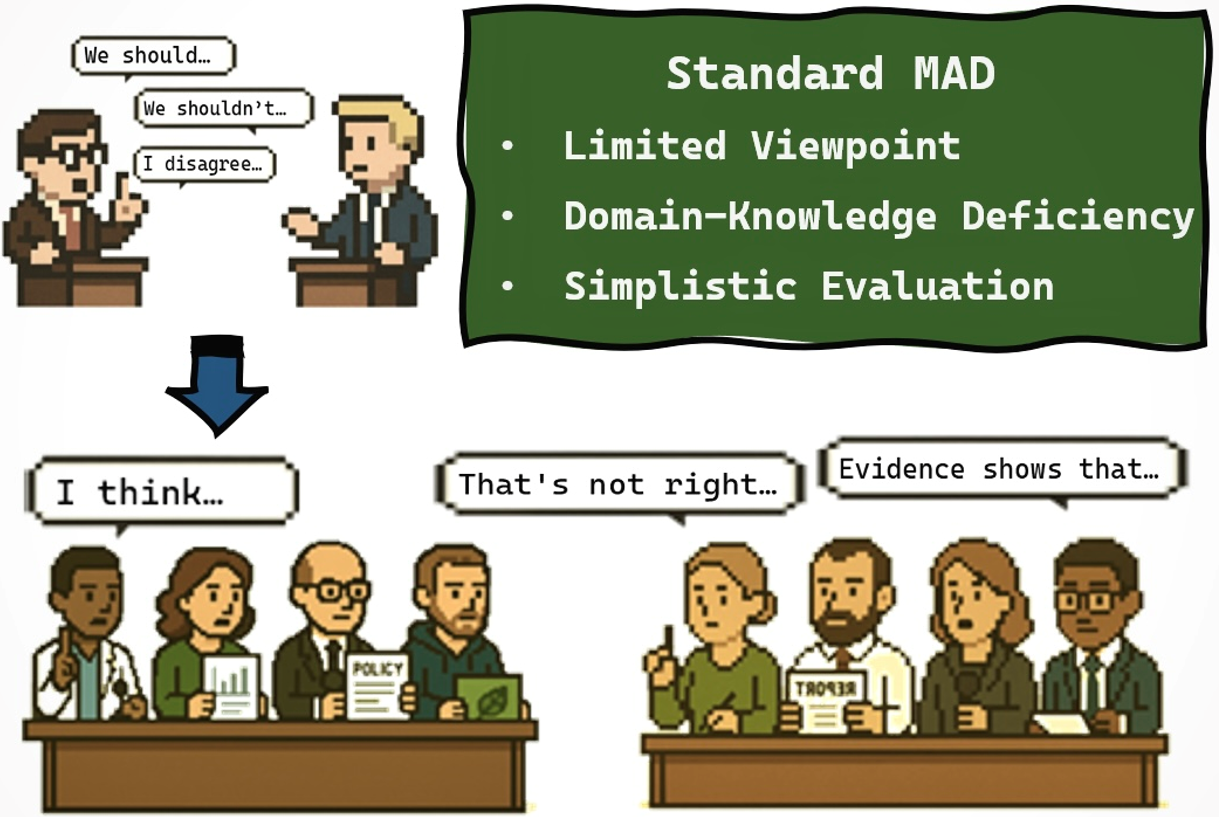}
  \caption{In Standard Multi-Agent Debate (SMAD), two debater agents participate in multi-turn exchanges, while a single judge agent evaluates the process. While effective for basic reasoning, it limits perspective diversity, lacks domain-specific expertise, and simplifies the evaluation. In contrast, D2D uses domain-specific agents with diverse viewpoints, allowing for deeper and more realistic argument exploration.}
  \label{figure1}
\end{figure}

To overcome the constraints, researchers have introduced multi‑step reasoning and multi‑agent strategies, including Chain‑of‑Thought (CoT) \citep{Weicot},
Self‑Reflection \citep{madaan2023selfrefine,ShinnReflexion}, and Multi‑Agent Debate (MAD) \citep{Du2024,liang-etal-2024-encouraging,li-etal-2024-improving-multi,amayuelas-etal-2024-multiagent}. While these methods have shown efficacy in mitigating hallucinations and enhancing reasoning ability, their evaluations are often restricted to controlled settings with limited contextual diversity, failing to capture the complexity of real-world misinformation \citep{DengAgents2025}. Moreover, existing MAD frameworks lack the structured process of fact-checking, where claims are systematically examined through evidence collection, counterargument analysis, and multi-dimensional evaluation conducted by domain experts \citep{slonim2021autonomous, masterman2024}. Current MAD frameworks focus on fragmented elements, employ generic agents, and neglect distinct debate stages, resulting in simplified binary judegments.

Inspired by the idea that “truth becomes clearer through debate,” we propose Debate-to-Detect (D2D), a novel MAD framework that simulates the fact-checking process through structured adversarial debates with LLM agents. Given an input text, D2D (i) identifies its topical domain, (ii) assigns each agent a concise domain profile, and (iii) orchestrates a five-stage debate comprising opening statements, rebuttals, free debate, closing statements, and judgment. A judging panel then evaluates the debate across five independent dimensions, producing an authenticity score that reflects both the truthfulness of the claim and the quality of the reasoning process. By reformulating misinformation detection as debate, D2D achieves higher accuracy while enhancing interpretability, aligning with real-world fact-checking practices.

Our contributions are summarized as follows: 

\textbf{(1) We introduce D2D, a structured deliberative framework for misinformation detection inspired by real-world fact-checking workflows.} D2D assigns domain-specific profiles to agents, engaging them in a five-stage progressive debate. This structured debate enhances logical coherence and facilitates stepwise evidence refinement, reflecting human reasoning patterns. Experiment results demonstrate that D2D not only significantly outperforms baseline methods but also remains robust on recently published news beyond GPT-4o’s pre-training.

\textbf{(2) We propose a multi-dimensional evaluation mechanism that redefines verdict generation in LLM-based misinformation detection.} Our design introduces a structured rubric comprising five dimensions: Factuality, Source Reliability, Reasoning Quality, Clarity, and Ethics. This schema enables D2D to produce interpretable authenticity scores with explicit rationale, reflecting rubric-based judgement practices in human debate.

\textbf{(3) We conduct a comprehensive analysis of the debate mechanism to examine how key components enhance misinformation detection.} Ablation studies underscore the complementary roles of domain profiles, stage design, and multi-dimensional evaluation. Stage-wise substitution further shows that debate phases differ in their demands on model capacity, with the judgement stage being most critical. Robustness tests confirm D2D’s resistance to biases such as speaker order and lexical framing. These results advance the understanding of multi-agent debate and support the design of more interpretable and reliable detection systems.

\begin{figure*}[t]
\centering 
  \includegraphics[width=0.79\linewidth]{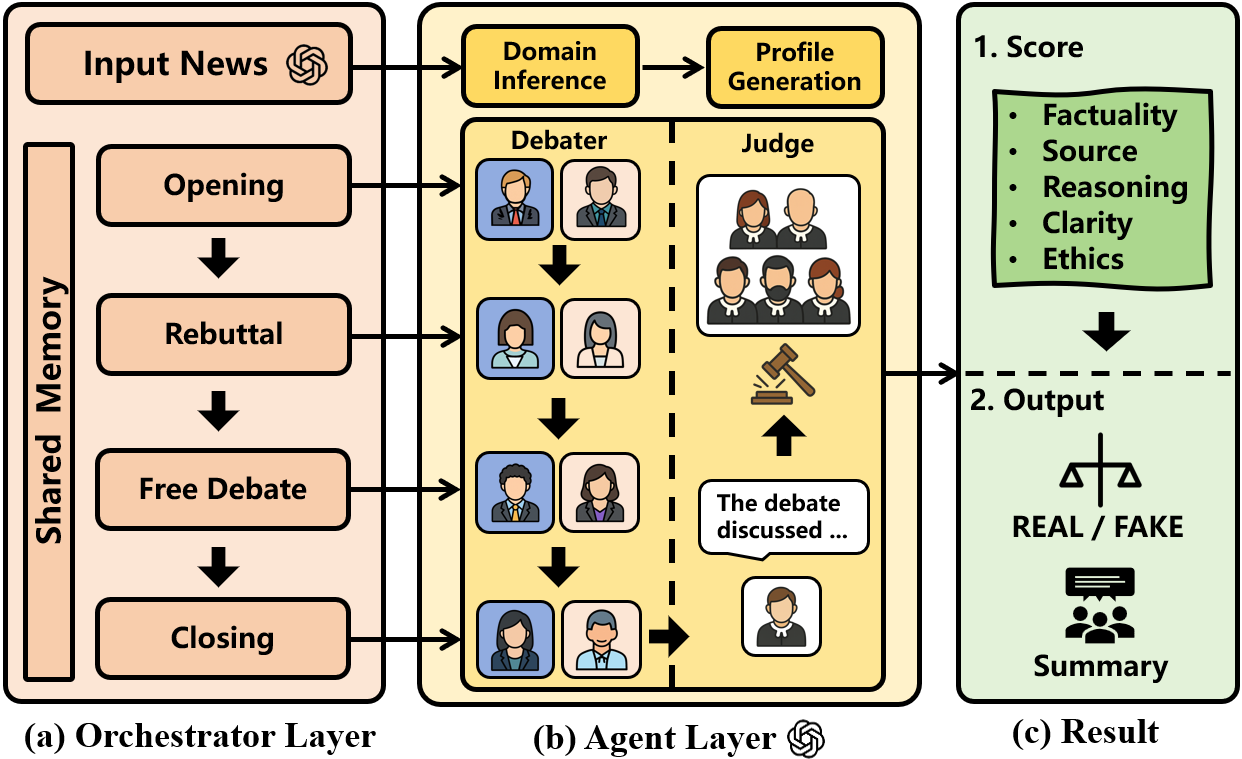}
  \caption {The D2D framework structures misinformation detection as a multi-agent debate, comprising two layers: \textbf{the Agent Layer and the Orchestrator Layer}. The Orchestrator Layer (a) coordinates the debate process through five stages—Opening, Rebuttal, Free Debate, Closing, and Judgement—while maintaining a shared memory. The Agent Layer (b) comprises domain-specific agents, including the Affirmative, Negative, and Judge roles.
  The Judge Agents evaluate the debate along five independent dimensions, and generates both a binary authenticity judgment and a debate summary.}
  \label{figure2}
\end{figure*}

\section{Related Work}
\subsection{Misinformation Detection}
The proliferation of misinformation across digital platforms has motivated extensive research on automated detection methods. Most existing approaches follow content-based paradigms, leveraging deep learning models to learn associations between textual features and veracity labels \citep{Nan2021,Mridha2021,Xu2024}. These methods incorporate lexical semantics, syntactic structure, and sentiment to build classifiers for misinformation detection. However, they often struggle with contextual understanding, particularly in complex or adversarial scenarios.

The emergence of LLMs has introduced new possibilities for misinformation detection \citep{liu-etal-2024-decoding,sharma2024fakenews}. Recent LLM-based misinformation detection incorporates synthetic data generation, multi-perspective reasoning, and instruction-based veracity assessment to enhance robustness and generalization \citep{He2023,wan-etal-2024-dell}. This transition facilitates more interpretable and scalable misinformation detection, particularly in zero-shot setting. However, most existing LLM-based misinformation detection methods still rely on a single agent, limiting their ability to capture the complexity of real-world cases. This limitation motivates the development of multi-agent approaches.

\subsection{Multi-Agent Debates}
Multi-Agent Debate (MAD) framework simulates a deliberative process in which multiple LLM-based agents interact iteratively to assess claims, challenge assumptions, and refine reasoning \citep{Du2024}. By distributing reasoning across agents with different roles or prompts, MAD better reflects the dynamic process of human argumentation and consensus building \citep{he-etal-2024-agentscourt}. Agents exchange arguments, rebuttals, and evaluations across multiple rounds, encouraging diverse reasoning paths and reducing the risk of early convergence \citep{liang-etal-2024-encouraging}. Prior work on MAD has examined various design choices, such as role assignment \citep{he-etal-2024-agentscourt}, communication structure \citep{li-etal-2024-improving-multi,amayuelas-etal-2024-multiagent}, and judgement aggregation \citep{park-etal-2024-predict}. Although these methods have proven effective in enhancing reasoning depth and diversity across different tasks, their application to misinformation detection remains largely unexplored.

Another limitation of existing MAD frameworks is their inability to capture the structured progress of real-world debates. Human deliberation is typically organized in distinct stages, each serving a specific purpose and contributing to progressive reasoning \citep{slonim2021autonomous,zhang2024}. In contrast, most MAD systems homogenize each interaction round, failing to differentiate between the stages \citep{cemri2025multiagentllmsystemsfail}. The lack of structural variation constrains their capacity to capture the dynamics of persuasion and rebuttal, which are crucial for robust misinformation detection.

\section{Our Framework: Debate to Detect}
Figure~\ref{figure2} illustrates the framework of \textbf{D2D}, consisting of two layers: \textbf{the Agent Layer}, which assigns role profiles and allocates tasks to enable diverse argumentation; and \textbf{the Orchestrator Layer}, which controls the debate flow and integrates judgments.

\subsection{Agent Layer}
The Agent Layer consists of three distinct roles: \textbf{Affirmative, Negative}, and \textbf{Judge}. The Affirmative and Negative sides each include four debater agents with a fixed stance of "\textbf{The Claim is Real}" or "\textbf{Fake}." This configuration follows the "tit for tat" strategy proposed by \citet{liang-etal-2024-encouraging}, encouraging diverse reasoning paths and reducing confirmation bias. Agent profiles are dynamically generated based on the topical domain of the input, ensuring context-aware argumentation.

To enable multi-dimensional evaluation and minimize bias, six judge agents are deployed, each evaluating arguments along specific dimensions. Unlike single-agent evaluations, the multi-judge setup enhances robustness and aligns with the ChatEval strategy for diversified assessment \citep{chan2024chateval}. Role-specific profiles further promote argumentative diversity and relieve the "Degeneration-of-Thought" (DoT) issue observed in LLM-based debates \citep{Du2024}.

\subsection{Orchestrator Layer}
The Orchestrator Layer structures the debate into five progressive stages: Opening Statement, Rebuttal, Free Debate, Closing Statement, and Judgement. This progression ensures both breadth and depth in reasoning. The \textbf{Opening Statement} introduces the core arguments from both sides, followed by the \textbf{Rebuttal}, where opposing claims are directly challenged. The \textbf{Free Debate} stage enables agents to flexibly extend, refine, and contest arguments, allowing the debate to evolve beyond scripted exchanges and surface novel reasoning paths. The process concludes with the \textbf{Closing Statement}, which consolidates arguments, and the \textbf{Judgement}, where the multi-judge panel delivers an evidence-based decision.

A key mechanism supporting this structure is the \textbf{Shared Memory}, which accumulates all prior arguments and evidence. Before each turn, the active agent receives a \textbf{compressed summary} of this memory. The summarization constrains generation by highlighting salient arguments, suppressing redundancy, and preserving coherence across stages. This design prevents drift and ensures that agents consistently engage with the central points of contention rather than digressing into irrelevant or repetitive content.

\subsection{Scoring Mechanism}

Following the Closing Statement, the Agent Layer will initiate a two-step judgement process: \textbf{(1) Neutral Synopsis:} A judge agent generates a comprehensive summary of the debate; \textbf{(2) Scoring:} Five independent judge agent assess both sides across the following dimensions \citep{SOPRANO2021102710}: Factuality, Source Reliability, Reasoning Quality, Clarity, and Ethics. Each Judge assigns complementary integer scores summing to 7 (e.g., 4:3, 5:2, 6:1), adhering to a strict zero-sum structure. This design guarantees an unambiguous outcome—since the total score across all dimensions is inherently imbalanced, a tie is mathematically impossible. Consequently, each news is definitively classified as REAL or FAKE.

\section{Experiment}
\subsection{Experimental Setup}

\textbf{Datasets.} We conduct experiments on two public datasets: Weibo21 \citep{Nan2021} and the FakeNewsDataset (consisting of FakeNewsAMT and Celebrity) \citep{perez-rosas-etal-2018-automatic}. To minimize interference from excessively long texts, the top 5\% of the longest samples are excluded. Additionally, the original Weibo21 dataset contains many low-quality samples that are are ambiguous or unverifiable, and we remove such samples to aovid the issue. The statistics of the preprocessed datasets are summarized in Table~\ref{table1}. We also report results on the original datasets and the error analysis in Appendix~\ref{why}.

\begin{table}[h!]
    \centering
    \fontsize{10pt}{12pt}\selectfont  
    \begin{tabularx}{\linewidth}{>{\raggedright\arraybackslash}Xccc}
        \toprule
        \textbf{Dataset} & \textbf{Fake} & \textbf{Real} & \textbf{Average Words} \\ 
        \midrule
        Weibo21          & 2373          & 2461          & 100.44                 \\
        FakeNewsDataset  & 466           & 466           & 211.73                 \\
        \bottomrule
    \end{tabularx}
    \caption{Statistics of two datasets}
    \label{table1}
\end{table}

\begin{table*}[ht!]
\centering
\setlength{\tabcolsep}{5pt} 
\renewcommand{\arraystretch}{1.12}
\begin{tabularx}{\textwidth}{
    >{\centering\arraybackslash}m{2cm}   
    *{4}{>{\centering\arraybackslash}X} |  
    *{4}{>{\centering\arraybackslash}X}    
}
\toprule
\multirow{2}{*}{\textbf{Method}} &
\multicolumn{4}{c|}{\textbf{Weibo21}} &
\multicolumn{4}{c}{\textbf{FakeNewsDataset}} \\
\cmidrule(lr){2-5}\cmidrule(lr){6-9}
& Accuracy & Precision & Recall & F1
& Accuracy & Precision & Recall & F1 \\ \midrule
BERT        & 75.64 & 78.50 & 77.06 & 77.77 & 77.30 & 77.60 & 78.33 & 77.96\\
RoBERTa     & 79.82 & 80.42 & 80.75 & 80.58 & 80.17 & 81.03 & 80.39 & 80.71\\
ZS         & 67.11 & 65.74 & 68.90 & 67.28 & 66.31 & 65.57 & 68.67 & 67.09 \\
CoT        & 74.04 & 72.74 & 75.35 & 74.02 & 72.32 & 71.14 & 75.11 & 73.07 \\
SR         & 76.33 & 75.68 & 76.32 & 76.00 & 73.71 & 74.29 & 72.53 & 73.40 \\
SMAD       & 77.02 & 76.76 & 76.27 & 76.52 & 74.79 & 74.42 & 75.54 & 74.97 \\
D2D w/o DP & 79.38 & 79.76 & 77.71 & 78.72 & 78.54 & 78.79 & 78.11 & 78.45 \\
D2D w/o SD & 80.33 & 79.90 & 80.07 & 79.98 & 78.33 & 77.73 & 79.40 & 78.56 \\
D2D w/o MJ & 78.88 & 78.51 & 78.21 & 78.36 & 76.72 & 76.42 & 77.25 & 76.84 \\
\textbf{D2D}        & \textbf{82.17} & \textbf{81.39} & \textbf{82.55} & \textbf{81.97} & \textbf{81.65} & \textbf{80.67} & \textbf{83.26} & \textbf{81.94} \\ \bottomrule
\end{tabularx}
\caption{Overall accuracy, precision, recall, and F1-score (\%) on \textit{Weibo21} and \textit{FakeNewsDataset}. D2D achieves the highest performance across all metrics, highlighting the impact of iterative reasoning, debate structure, and evaluation design.}
\label{table2}
\end{table*}

\textbf{Baselines.} We compare our D2D framework with the following baselines:
\begin{itemize}
    \item \textbf{BERT} \cite{devlin-etal-2019-bert}: A fine-tuned BERT-base model for binary classification.
    \item \textbf{RoBERTa} \cite{liu2019robertarobustlyoptimizedbert}: A fine-tuned RoBERTa-base model with the same setup as BERT, serving as a stronger discriminative baseline.
    \item \textbf{Zero-Shot (ZS):} A single LLM performs direct classification of each news item without other prompting.
    \item \textbf{Chain-of-Thought (CoT)}\citep{Weicot}: The model generates an explicit step-by-step reasoning process before producing the final prediction.
    \item \textbf{Self-Reflect (SR) }\citep{madaan2023selfrefine}: The model iteratively critiques and revises its own outputs until the self‑evaluation indicates convergence or no further improvement.
    \item \textbf{Standard Multi-Agent Debate (SMAD):} Two debater agents with generic profiles engage in a fixed number of debate rounds (set to four here for alignment with our framework). A single judge agent evaluates the debate and make the judgement.
\end{itemize}

\textbf{D2D Variants}. To evaluate the impact of different components in the D2D framework, we design three ablated versions:
\begin{itemize}
\item\textbf{D2D w/o DP (Domain Profile):} This variant removes domain-specific profiles, replacing them with a generic profile for all participants to assess the influence of domain knowledge.

\item\textbf{D2D w/o SD (Stage Design):} This variant eliminates the structured four-stage debate process, replacing it with a continuous four-round discussion where agents interact without predefined roles or prompt-specific duties.

\item\textbf{D2D w/o MJ (Multi-dimensional Judgement):} This variant eliminates the multi-dimensional judgement mechanism, and a single-dimensional judgement is applied, focusing on the factuality of claims.

\end{itemize}

\textbf{Model Configuration.} All experiments use GPT-4o as the base model. LLM-agents are initialized with predefined prompts provided in Appendix~\ref{prompt}. Agent response lengths are capped at 1024 tokens. Domain inference and final judgment are conducted with a temperature of 0.0 to ensure stability. To encourage diversity, profile generation and debate responses across all stages use a temperature of 0.7. Unless otherwise specified, the number of Free Debate rounds is fixed at 1.

\begin{figure*}[t]
\centering 
  \includegraphics[width=0.85\linewidth]{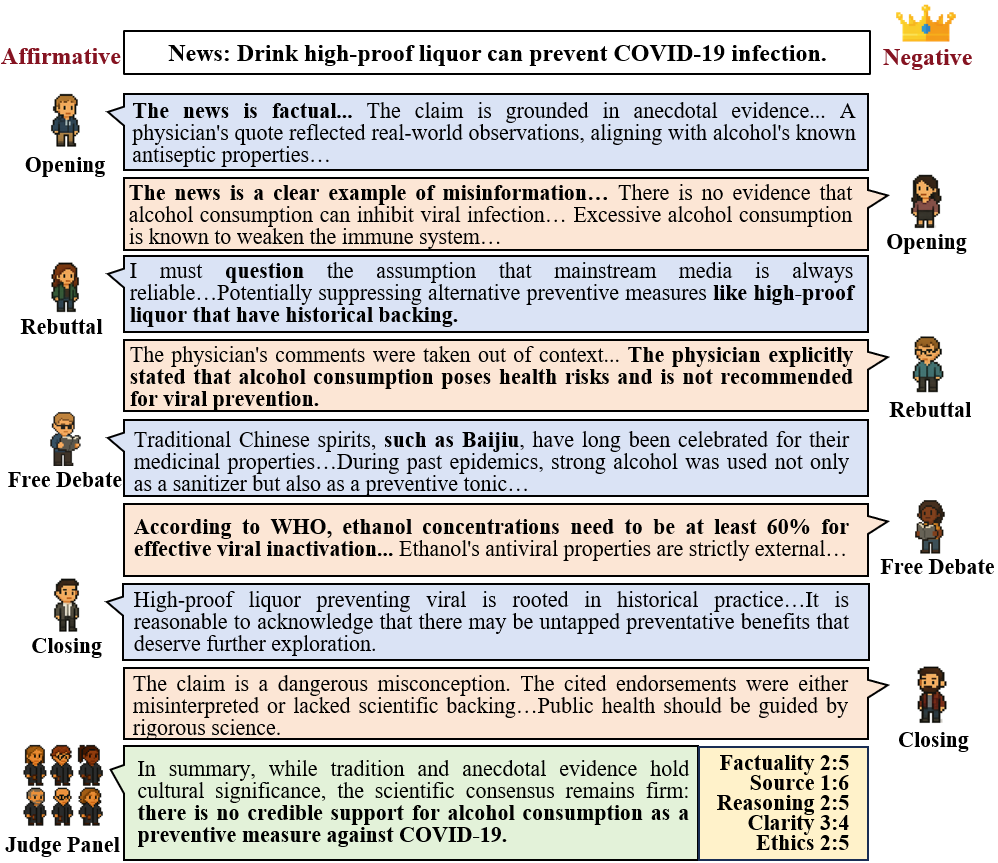}
\caption{
Case Study – A Demonstration of the Structured MAD in the D2D Framework. The process reflects realistic argumentative strategies, including rhetorical misinformation tactics and factual rebuttals, while progressively refining evidence through agent interaction.}

  \label{figure3}
\end{figure*}

\subsection{Results}
We measure the performance using four standard metrics: accuracy, precision, recall, and F1-score. Table~\ref{table2} presents the overall results of D2D, baselines and ablated variants on the two datasets. Across all datasets and metrics, D2D achieves the best performance, significantly outperforming both fine-tuned transformers and prompting-based approaches. Although models like RoBERTa achieve competitive performance, they remain limited in interpretability.

The improvement from ZS to CoT and SR demonstrates a clear improvement in misinformation detection, highlighting the benefits of iterative reasoning mechanisms. Specifically, CoT enhances performance over ZS by approximately 6.74\% and 5.98\% in F1-score on Weibo21 and FakeNewsDataset, respectively. The SR method further refines these results, achieving 76.00\% and 73.40\% in F1-score on the two datasets, reflecting the effectiveness of self-evaluation and iterative refinement. Incorporating adversarial interactions through SMAD results in additional gains, with 76.52\% and 74.97\% F1-score on Weibo21 and FakeNewsDataset, respectively, representing a small improvement over SR, indicating that structured two-agent debate enhances evidence evaluation by introducing conflicting perspectives.

D2D achieves the highest performance across all metrics on both datasets, with 81.97\% and 81.94\% F1-score on Weibo21 and FakeNewsDataset, respectively. Ablation studies reveal that removing Domain Profiles leads to F1-score reductions of 3.25\% on Weibo21 and 3.49\% on FakeNewsDataset, closely aligning with D2D's gains over SMAD. The removal of Stage Design results in smaller declines of 1.99\% on Weibo21 and 3.38\% on FakeNewsDataset, highlighting the structured stages' role in enhancing logical coherence and handling longer texts. Furthermore, eliminating the Multi-Dimensional Judgement mechanism causes more pronounced drops of 3.61\% on Weibo21 and 5.10\% on FakeNewsDataset, underscoring its critical contribution to the assessment. These results emphasize the synergistic effects of domain-specific profiling, structured debate stages, and multi-dimensional evaluation in optimizing the judgement reliability.

\subsection{Case Study}
Figure~\ref{figure3} presents a representative debate example within the D2D framework, focusing on the claim “drink high-proof liquor can prevent COVID-19 infection” from Weibo21. We highlight three observations that illustrate how D2D reflects the patterns of realistic argumentation while enhancing factual resolution.

\textbf{(1)	Stage coherence.}

The framework begins by assigning concise health-related profiles to all debater and judge agents. Both sides adhere to the five-stage structure. In the Opening Statement, the Affirmative introduces anecdotal evidence and a misquoted physician statement, whereas the Negative contextualizes the argument with epidemiological reasoning. In the Rebuttal stage, the Negative systematically refutes the cited endorsement by referencing the original interview, and the Affirmative counters by questioning the credibility of mainstream media, a rhetorical strategy frequently observed in real-world misinformation discourse.

\textbf{(2)	Progressive evidence refinement.}

The dialogue demonstrates incremental evidence development. The Affirmative cites traditional Chinese spirits as an example, and the Negative introduces WHO-published ethanol-inactivation thresholds as a counter. This exchange demonstrates that agents are not merely repeating predefined outputs but dynamically revising their claims in response to new information, exhibiting the adaptive reasoning behavior that the D2D framework is designed to facilitate.

\textbf{(3) Criterion-Based Evaluation.}

Following the Closing statements, one Judge provides a neutral summary of the debate, while the remaining five assign scores across predefined evaluation dimensions. Accuracy (2:5), Source Reliability (1:6), Reasoning (2:5) and Ethics (2:5) overwhelmingly favor the Negative, while Clarity shows a tighter score gap of 3:4, reflecting the Affirmative’s stylistic appeal despite weak factual grounding. The final aggregate (10:25) results in a clear FAKE classification.

This case shows that D2D can provide accurate judgements through structured dialogue and stepwise evidence exchange. The debate process reflects real-world argumentative patterns and provides clear, interpretable justifications results.

\begin{table*}[]
\centering
\begin{tabular}{cccccc}
\toprule
\multirow{2}{*}{\textbf{Perturbation}} & \multirow{2}{*}{\textbf{Judgement Result}} & \multicolumn{2}{c}{\textbf{Fake}} & \multicolumn{2}{c}{\textbf{Real}} \\ \cmidrule(lr){3-4} \cmidrule(lr){5-6}
                                       &                                           & \(\Delta \leq 5\)  & \(5 \leq \Delta \leq 10\)  & \(\Delta \leq 5\)  & \(5 \leq \Delta \leq 10\)  \\ \midrule
\multirow{2}{*}{Speaking Order}         & Consistent                                & 90            & 3                 & 93            & 2                 \\
                                       & Inconsistent                              & 7             & 0                 & 5             & 0                 \\
\multirow{2}{*}{Neutral   Relabeling}  & Consistent                                & 94            & 1                 & 96            & 0                 \\
                                       & Inconsistent                              & 5             & 0                 & 4             & 0                 \\ 
\bottomrule
\end{tabular}
\caption{Robustness of D2D to Speaker Order and Lexical Framing Perturbations. Over 90\% of the samples demonstrate strong robustness ($\Delta \leq 5$), indicating that D2D is highly resilient to biases arising from speaker order and lexical framing variations.}
\label{table3}
\end{table*}

\section{Analysis}

\subsection{Which Debate Stage Matters Most? }
\label{sec5.1}
In classical debate theory, each stage serves a distinct rhetorical function: Opening establishes the argument, Rebuttal introduces the counterpoints, Free Debate facilitates interactive reasoning, Closing consolidates key arguments, and Judgement delivers the final evaluation. To quantify the relative contribution of each stage and examine how model capability affects performance, we conduct a controlled cross-model substitution experiment on the FakeNewsDataset. Specifically, the model at each stage is replaced with either weaker GPT-3.5-turbo or stronger GPT-4.1, while keeping the remaining stages unchanged.

\begin{figure}[h!]
\centering
  \includegraphics[width=0.95\linewidth]{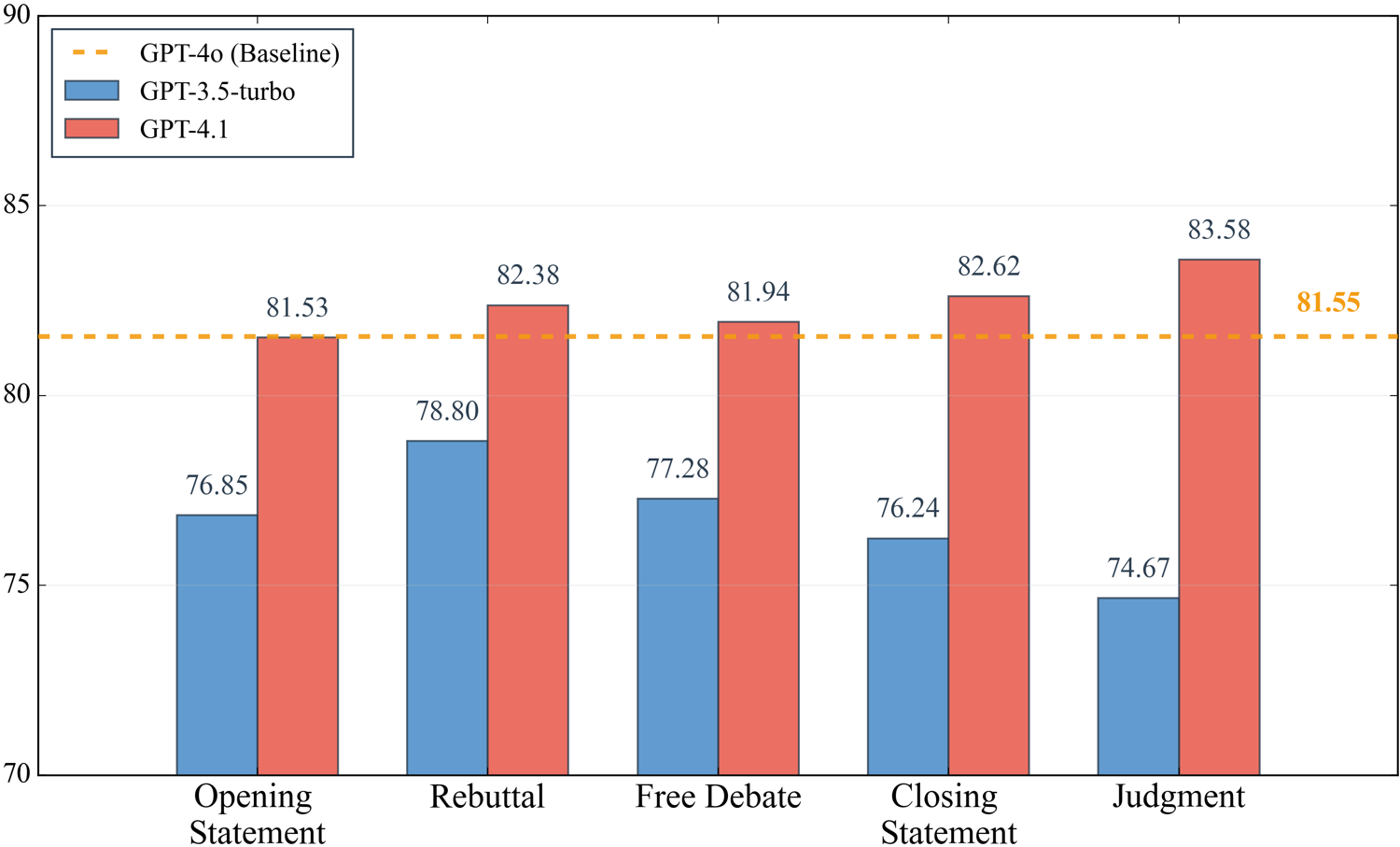}
  \caption {Performance Comparison of Model Variants Across Debate Stages in the D2D.}
  \label{figure4}
\end{figure}

Figure~\ref{figure4} presents the F1-score for each configuration. Compared to the GPT-4o baseline (81.55\%), substituting GPT-4.1 consistently improves performance across all stages, with the most substantial gain observed in the Judgement stage (+3.03\%). Meanwhile, replacing GPT-3.5-turbo leads to performance drops, with the largest drop also occurring at the Judgement (\textminus{}6.87\%). These findings align with prior research by \citet{liang-etal-2024-encouraging}, which similarly identifies the Judgement stage as the most critical component in MAD frameworks.

\subsection{Do Speaker Order and Side Labels Influence Judgements?}
\label{sec5.2}

LLMs are known to exhibit biases associated with speaker order and lexical framing, potentially influencing outputs in adversarial dialogue settings by favoring the side that speaks first or carries a more positively connoted label \citep{Sultan2024,wang2025large}. To evaluate whether such biases affect the fairness of D2D, we design two controlled perturbation experiments targeting speakeing order and side labels, respectively. 

Specifically, we randomly select 100 fake and 100 real samples from the FakeNewsDataset and evaluate the robustness of D2D by measuring (i) judgement consistency and (ii) the distribution of score deviations under the 35-point scale. The absolute difference in judgement scores, denoted as $\Delta$, serves as a key measure of consistency across perturbations. It captures the deviation of judgement scores between the original and perturbed configurations. $\Delta \leq 5$ indicates strong consistency, while $5 < \Delta \leq 10$ suggests moderate variation. Table~\ref{table3} presents the results.

\textbf{(a) Speaking Order Permutation.}

In this experiment, the initial speaking order of the Affirmative and Negative sides are reversed, keeping all other components constant. Among FAKE samples, 90 samples remain consistent within a 5-point deviation, with an additional 3 within a 10-point deviation. Only 7 cases show variations, all within 5 points. For REAL samples, 93 samples stay within the 5-point deviation, while the remaining 5 disagreements also fall within 5 points. These results suggest that D2D is robust to order-based biases.

\textbf{(b) Neutral Relabeling.}

To evaluate the susceptibility to lexical framing, we replace the terms "Affirmative" and "Negative" with neutral terms: "Supporter" and "Skeptic" in all prompts. For FAKE samples, 94 instances remained within a 5-point range, with 1 more case within 10 points. Verdict inconsistencies were minimal (5 for FAKE, 4 for REAL), all within 5 points. The result demonstrates D2D's robustness to lexical framing effects.

\subsection{The Influence of Debate Rounds}
\label{sec5.3}

The number of debate rounds in MAD have been shown to significantly impact the performance of reasoning tasks\citep{liang-etal-2024-encouraging,Du2024}. To further explore the adaptability of D2D, we conduct experiments on the FakeNewsDataset, stratified by text length and varied the number of debate rounds from 1 to 6. The rounds configurations are shown in Table~\ref{table4}:

\begin{table}[h!]
\centering
\small
\begin{tabular}{c|l}
\toprule
\textbf{Rounds} & \textbf{Included Debate Stages} \\
\midrule
1 & Opening only \\
2 & Opening+Closing \\
3 & Opening+Rebuttal+Closing \\
4 & Opening+Rebuttal+Free Debate+Closing \\
5 & Opening+Rebuttal+2×Free Debate+Closing \\
6 & Opening+Rebuttal+3×Free Debate+Closing \\
\bottomrule
\end{tabular}
\caption{Debate Stage Configurations for Different Round Settings}
\label{table4}
\end{table}

\begin{figure}[h!]
\centering
  \includegraphics[width=0.95\linewidth]{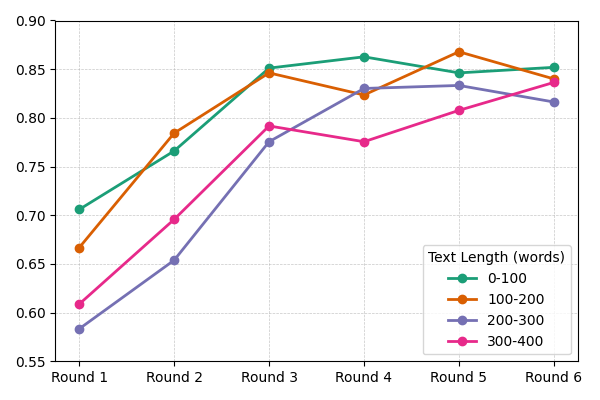}
  \caption {Effect of Debate Rounds on F1-Score Across Different Text Length Intervals}
  \label{figure5}
\end{figure}

We select 50 samples from four text length range (0-100 words, 100-200 words, 200-300 words, and 300-400 words), ensuring a balanced representation of fake and real samples (25 fake, 25 real) in each group. Figure~\ref{figure5} presents the performance F1-Score across the different configurations.

The results reveal that the effectiveness of debate rounds is significantly influenced by text length. For shorter texts (0–100 words), the optimal configuration is observed at 4 rounds; For slightly longer texts (100–200 words), the optimal is achieved at 5 rounds, suggesting that additional debate iterations contribute to argument refinement and correction.

For medium-length texts (200–300 words), the optimal performance is also observed at 5 rounds. This demonstrates that deeper rounds provide more comprehensive exploration of reasoning paths, enhancing judgement accuracy; For longer texts (300–400 words), the highest performance is achieved at 6 rounds, reflecting the need for extended deliberation to navigate complex narratives effectively.

These observations align with the findings of \citet{liang-etal-2024-encouraging} and \citet{li-etal-2024-improving-multi}, which highlight the importance of iterative reasoning stages in reducing information overload for shorter texts while enhancing argument development for longer claims. Meanwhile, the results also indicate that exceeding the optimal number of rounds can have negative effects, particularly for shorter texts where additional rounds fail to provide further improvements.

\subsection{Generalization to Latest Published News}
\label{sec5.4}

One common critique of LLM-based detectors is their potential reliance on memorized content from pre-training corpora \citep{das2025fakenewsdetectionllm}. To evaluate the generalization capability of D2D beyond pre-trained knowledge, we construct a benchmark consisting of 596 Chinese news samples (342 real, 254 fake) sourced from the Chinese Internet Rumor Dispelling Platform\footnote{\url{www.piyao.org.cn}} between January and April 2025—a period postdating the GPT-4o pre-training cut-off in June 2024.

\begin{table}[h!]
    \centering
    \begin{tabular}{lcc}
        \toprule
        \textbf{Method} & \textbf{Accuracy} & \textbf{F1} \\ 
        \midrule
        ZS   & 74.50 & 68.46 \\ 
        SMAD & 78.69 & 73.92 \\ 
        \textbf{D2D}  & \textbf{83.92} & \textbf{79.83} \\ 
        \bottomrule
    \end{tabular}
    \caption{Accuracy and F1-score (\%) on the Latest News, and D2D achieves the highest performance across both metrics.}
    \label{table5}
\end{table}

As shown in Table~\ref{table5}, D2D achieves an accuracy of 83.92\% and an F1-score of 79.83\%, significantly outperforming SMAD, which attains 78.69\% accuracy and 73.92\% F1-score, as well as the zero-shot GPT-4o baseline. A manual inspection of the 254 fake samples confirm the absence of verbatim overlaps with publicly indexed sources prior to June 2024, indicating that D2D is not merely retrieving memorized content.

\section{Conclusion}

In this paper, we introduced Debate-to-Detect (D2D), a structured multi-agent debate framework that reformulates misinformation detection as an adversarial deliberation process. By assigning domain-specific profiles, orchestrating a five-stage debate, and applying a multi-dimensional evaluation rubric, D2D improves both accuracy and interpretability over strong baselines. Our analysis further demonstrates its robustness, generalization beyond memorized content, and resilience to biases such as speaker order and lexical framing. A case study illustrates D2D’s ability to progressively refine evidence and deliver criterion-based evaluations, closely mirroring real-world fact-checking workflows.

Future work will focus on extending D2D to multimodal misinformation (e.g., images, videos, and deepfakes), integrating external fact-checking databases to reduce hallucinations, and enhancing its persuasive capacity—not only classifying claims as true or false, but also explaining why they are misleading. These directions are essential for advancing reliable, transparent, and socially responsible AI systems for misinformation detection.

\section*{Limitations}
\textbf{Interaction Cost.} D2D involves 5 debate stages and the coordination among 14 agents, resulting in considerable computational. To enable deployment in real-time settings, such as social media monitoring, future work may be expected to explore adaptive truncation strategies or lightweight models that maintain diversity without compromising quality.

\noindent
\textbf{Evidence Modality.} Currently, D2D operates on textual input and does not incorporate external links, images, or videos, and thus lacks the capacity to detect multimodal misinformation such as deepfakes. Future work will focus on extending D2D's reasoning capabilities to encompass multimodal evidence, enabling more comprehensive misinformation detection.

\noindent
\textbf{Scalability and Real-time Adaptation.} The performance of D2D is inherently tied to the capabilities of the underlying LLMs. Any deficiencies or biases in the LLM's pre-trained knowledge can propagate through the framework, affecting judgment reliability. This dependency introduces vulnerabilities, particularly when encountering domain-specific misinformation where LLM knowledge is outdated. Future work should consider integrating external knowledge bases, such as fact-checking repositories and domain-specific databases, to enhance real-time accuracy and reduce reliance on LLM-generated assumptions.

\section*{Ethics Statement}

A major concern in LLM-based misinformation detection is the risk of biased or erroneous inferences, arising from data imbalances or hallucinations. Moreover, D2D’s agent-driven debates, while designed to simulate human argumentation, may fall short in capturing the nuance required for real-world fact-checking, particularly in culturally sensitive or politically charged contexts. These concerns are especially acute in high-stakes domains like law, medicine, and politics, where misleading arguments may erode trust, destabilize communities, or harm individual rights.

\bibliography{references}

\appendix
\section*{Appendix}

\section{Experiments on full datasets}
\label{why}

In this appendix, we present the experimental results on the original datasets. Our main experiments are conducted on preprocessed versions of Weibo21 and FakeNewsDataset to ensure data quality and reduce noise. By evaluating the raw datasets, we illustrate how low-quality and ambiguous samples can degrade the model performance. The statistics of original datasets are presents in Table~\ref{table6}. The preprocessed datasets are also available at \href{https://anonymous.4open.science/r/emnlp_d2d-36E2/}{\texttt{4open.science/emnlp\_d2d-36E2}}.

\begin{table}[h!]
    \centering
    \fontsize{10pt}{12pt}\selectfont  
    \begin{tabularx}{\linewidth}{>{\raggedright\arraybackslash}Xccc}
        \toprule
        \textbf{Dataset} & \textbf{Fake} & \textbf{Real} & \textbf{Average Words} \\ 
        \midrule
        Weibo21          & 2795          & 2956          & 92.08                 \\
        FakeNewsDataset  & 490           & 490           & 276.12                 \\
        \bottomrule
    \end{tabularx}
    \caption{Statistics of two original datasets.}
    \label{table6}
\end{table}

\subsection{Performance on Full Datasets}

Table~\ref{table7} presents the performance of the D2D framework on the original datasets. As observed, the results exhibit a decline compared to the preprocessed datasets, particularly in terms of Recall. This disparity indicates that a significant portion of fake samples remains undetected by the model, resulting in a substantial number of false negatives.

\begin{table*}[ht!]
\centering
\setlength{\tabcolsep}{5pt} 
\renewcommand{\arraystretch}{1.12}
\begin{tabularx}{\textwidth}{
    >{\centering\arraybackslash}m{2cm}   
    *{4}{>{\centering\arraybackslash}X} |  
    *{4}{>{\centering\arraybackslash}X}    
}
\toprule
\multirow{2}{*}{\textbf{Method}} &
\multicolumn{4}{c|}{\textbf{Weibo21}} &
\multicolumn{4}{c}{\textbf{FakeNewsDataset}} \\
\cmidrule(lr){2-5}\cmidrule(lr){6-9}
& Accuracy & Precision & Recall & F1
& Accuracy & Precision & Recall & F1 \\ \midrule
ZS         & 65.14 & 65.93 & 58.50 & 61.99 & 64.59 & 63.88 & 67.14 & 65.47\\
D2D        & 78.79 & 82.00 & 72.20 & 76.79 & 81.22 & 80.72 & 82.04 & 81.38 \\
\bottomrule
\end{tabularx}
\caption{Overall accuracy, precision, recall, and F1-score (\%) on original datasets.}
\label{table7}
\end{table*}

\subsection{Error Analysis}

\begin{figure*}[h!]
\centering
  \includegraphics[width=1\linewidth]{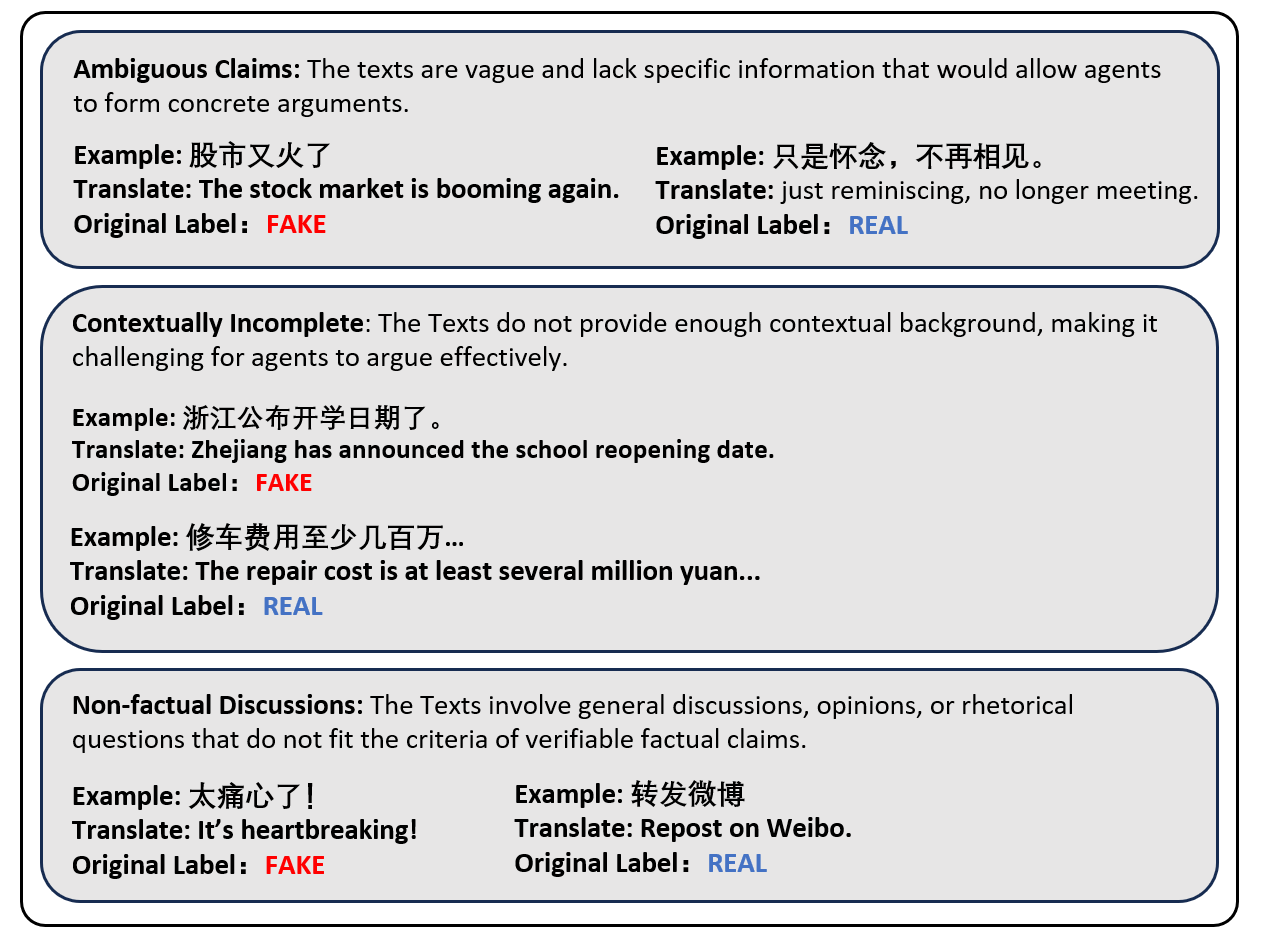}
  \caption {Examples of low-quality samples in Weibo21.}
  \label{figure6}
\end{figure*}

Upon analysis, a significant portion of the observed performance degradation can be attributed to low-quality samples, particularly in the Weibo21 dataset. These samples often exhibit poor structural coherence, substantial noise, or represent unverifiable claims that elude standard fact-checking procedures. We illustrate examples of these problematic samples in Figure~\ref{figure6}. Consequently, the preprocessing is not only beneficial but necessary for enhancing model interpretability and performance consistency.

\section{Prompts Archive}
\label{prompt}

\noindent
\textbf{Domain Inference:}

Classify the domain of the following claim in one or two words (e.g., politics, finance, sports, technology, health). Claim:\{input\}
\vspace{2mm}

\noindent
\textbf{Profile Generation:}

The domain is \{domain\}. Provide a brief professional profile (3-4 sentences) for a debater in \{stage\_name\} stage role relevant to this domain.
\vspace{2mm}

\textbf{Profile Example:}

\textbf{Health:} As an experienced health communications expert, I specialize in analyzing and disseminating medical news and information. With a decade of experience working alongside healthcare professionals, researchers, and policymakers, I have a profound understanding of the complexities and dynamics that shape the health news landscape. My work is grounded in ensuring that health news is conveyed accurately and responsibly, leveraging evidence-based research to inform the public. Passionate about promoting health literacy, I am dedicated to enhancing the quality and reliability of health-related news.

\textbf{Finance:} As a financial analyst with extensive experience in equity markets, I focus on evaluating corporate disclosures, earnings reports, and market signals. My expertise lies in assessing financial credibility and detecting inconsistencies across financial statements and media reports. Having collaborated with regulatory agencies and institutional investors, I bring a critical perspective to debates on corporate transparency. I am committed to ensuring that financial information is communicated with clarity and integrity.

\textbf{Environment:} As a climate policy researcher, I have worked with international organizations to assess the impact of environmental regulations on energy sectors. My expertise includes analyzing emission reduction policies, carbon trading mechanisms, and climate adaptation strategies. I bring an evidence-driven approach to discussions of environmental claims, ensuring alignment with the latest scientific findings and policy frameworks. My goal is to promote informed decision-making and constructive dialogue on sustainability issues.
\vspace{2mm}

\noindent
\textbf{Shared Memory:}

Given the following debate history: \{debate\_history\}

Summarize the key points from both the Affirmative and Negative sides, ensuring the following aspects are preserved:
1. The main claim and its justification.
2. Key arguments and supporting evidence from both sides.
3. Notable rebuttals and counterarguments.
4. Any unresolved contradictions or logical conflicts.

Your summary should be concise yet comprehensive, allowing future agents to understand the debate's progression without losing important context. Aim to reduce redundancy while maintaining logical coherence.
\vspace{2mm}

\noindent
\textbf{Opening Statement:}

\{Profile\}

The claim under discussion is: \{input\}. 
Your assigned stance is \{fixed\_stance\}. 

Based on your designated role and the available argument history, construct a well-structured opening statement that convincingly defends your stance. 
Make sure to employ logical reasoning, relevant evidence, and clear argumentation to support your position.
\vspace{2mm}

\noindent
\textbf{Rebuttal:}

\{Profile\}

The claim under discussion is: \{input\}. 
Your assigned stance is \{fixed\_stance\}.
The previous argument presented was: \{Shared\_Memory\}. 

Identify the key weaknesses or logical inconsistencies in the opponent's argument and provide a well-structured rebuttal. 
Leverage relevant evidence and logical reasoning to effectively counter the claims made. Aim to challenge the validity of the argument while reinforcing your own position.
\vspace{2mm}

\noindent
\textbf{Free Debate:}

\{Profile\}

The claim under discussion is: \{input\}. 
Your assigned stance is \{fixed\_stance\}.
The previous argument presented was: \{Shared\_Memory\}. 

Building on your previous arguments and responding to the latest claims, provide a well-structured continuation of the debate. 
Focus on addressing any unresolved contradictions, introducing new evidence if necessary, and strengthening your stance with logical reasoning.
\vspace{2mm}

\noindent
\textbf{Closing Statement:}

\{Profile\}

The claim under discussion is: \{input\}. 
Your assigned stance is \{fixed\_stance\}.
The final evaluation is approaching. The previous argument presented was: \{Shared\_Memory\}. 

Using this information, summarize your key arguments and highlight the most compelling evidence presented throughout the debate. 
Emphasize the logical coherence of your stance, address any lingering concerns or contradictions raised by the opposition, and consolidate your position. 
Conclude with a clear and decisive statement that reinforces your stance as the more rational and evidence-based perspective.

\vspace{2mm}

\noindent
\textbf{Judgement of Summary}

\{Profile\}

You are assigned the role of a Judge responsible for summarizing the key points presented during the debate. Your task is to produce a concise and neutral summary that accurately reflects the main arguments from both the Affirmative and Negative sides. 

The previous argument presented was: \{Shared\_Memory\}. 

Focus on the following aspects:

1. The main claim and its context.

2. Key supporting arguments presented by the Affirmative side.

3. Key counterarguments raised by the Negative side.

4. Notable rebuttals and their logical coherence.

5. Any unresolved contradictions or gaps in reasoning.
\vspace{2mm}

\noindent
\textbf{Judgement of Evaluation}

\{Profile\}

You are assigned the role of a Judge, responsible for evaluating the quality and validity of the arguments presented during the debate. Affirmatives defend the claim as factual, and Negatives argue that the claim is misleading or fake.

The previous argument presented was: \{Shared\_Memory\}. 

Your task is to assess the arguments from both the Affirmative and Negative sides based on the \{dimension\_name\} dimension. 

For this dimension, assign an integer score to each side based on how convincingly they support their position relative to the truth. The two scores must add up to exactly 7.

Return the following JSON format:\{Affirmative: X, Negative: Y\}.

\end{document}